# Doubly Attentive Transformer Machine Translation

**Hasan Sait Arslan**   **Mark Fishel**
Institute of Computer Science
University of Tartu, Estonia
{hasan90,fishel}@ut.ee

**Gholamreza Anbarjafari**
Institute of Technology
University of Tartu, Estonia
shb@ut.ee

## Abstract

In this paper a doubly attentive transformer machine translation model (DATNMT) is presented in which a doubly-attentive transformer decoder normally joins spatial visual features obtained via pretrained convolutional neural networks, conquering any gap between image captioning and translation. In this framework, the transformer decoder figures out how to take care of source-language words and parts of an image freely by methods for two separate attention components in an Enhanced Multi-Head Attention Layer of doubly attentive transformer, as it generates words in the target language. We find that the proposed model can effectively exploit not just the scarce multimodal machine translation data, but also large general-domain text-only machine translation corpora, or image-text image captioning corpora. The experimental results show that the proposed doubly-attentive transformer-decoder performs better than a single-decoder transformer model, and gives the state-of-the-art results in the English-German multimodal machine translation task.

## 1 Introduction

Neural Machine Translation (NMT) has been effectively handled as a sequence to sequence learning issue (Kalchbrenner and Blunsom, 2013; Cho et al., 2014; Sutskever et al., 2014), where each training example consists of one source and one target variable-length sequences, with no earlier data on the arrangement between the two.

With regards to NMT, (Bahdanau et al., 2014) investigated the fact that the use of a fixed length vector is a bottleneck in enhancing the execution of the essential encoder-decoder architecture in neural machine translation, and proposed to broaden this by enabling a model to naturally scan for parts of a source sentence that are pertinent to predicting an objective word, without forming these parts as a hard section expressly. Also (Xu et al., 2015) proposed an attention based model for the task of neural image captioning (NIC) where a model figures out how to take care of particular parts of an image representation as it creates its caption (target) in normal language. Similarly, (Calixto et al., 2017) proposed an attention based approach for the task of Multi-modal Neural Machine Translation (MNMT) where a model figures out when and where to attend to the image or the source text when generating the words in the target language.

Lately the attention mechanism has been used in addition to NMT also in NIC and MNMT. In this work, we propose an end-to-end attention-based MNMT approach based on the transformer model (Vaswani et al., 2017), which fuses together two independent attention vectors, one over source-language words and the other over various areas of an image.

Our main contributions are:

- We propose a novel transformer-attention-based MNMT which makes use of spatial visual features and transformer encoder output features.

- We additionally use more than 90 M text-only corpora from OPUS (Tiedemann and Nygaard, 2004), and image captioning dataset Flickr30k

(Plummer et al., 2015) to pretrain machine translation and image captioning models and show that our MNMT model can efficiently exploit them.

- We demonstrate that images bring helpful data into transformer NMT model, in circumstances in which sentences describe objects represented in the image.

The rest of this paper is organized as takes after. We first examine past related work in Section 2, then describe the related background information (Section 3) and introduce our approach (Section 4). In Section 5, we introduce the datasets we use for training and assessing our models, in Section 6 we talk about our exploratory setup and investigate and examine our outcomes. At last, in Section 7 we present conclusions and list a few directions for future work.

## 2 Related work

There has been a lot of work on natural language generation from non-textual inputs. (Mao et al., 2014) introduced a multimodal Recurrent Neural Network model for creating sentence depictions to clarify the substance of images. It specifically models the likelihood of creating a word given past words and the image. Image captions are produced by sampling from this distribution. The model comprises of two sub-networks: a deep recurrent neural network for sentences and a deep convolutional neural network for images. (Vinyals et al., 2014) introduced a generative model in view of a deep recurrent design that consolidates late advances in computer vision and machine translation and that can be used to create regular sentences depicting an image. The model is trained to augment the probability of the target description sentence given a training image. (Elliott et al., 2015) displayed a way to deal with multi-language image description uniting bits of knowledge from neural machine translation and neural image captioning. To make a description of an image for a given target language, their sequence generation models condition on feature vectors from the image, the depiction from the source language, as well as a multimodal vector computed over the image and a description in the source language. (Venugopalan et al., 2015) propose an end-to-end sequence-to-sequence model to create subtitles for video recordings. For this they use recurrent neural networks, particularly LSTMs. Their LSTM model is trained on video-sentence matches and figures out how to relate a sequence of video frames to a sequence of words so as to produce a depiction of the occasion in the video cut. Their model normally can take in the fleeting structure of the sequence of frames in addition the sequence model of the produced sentences. (Xu et al., 2015) presented an attention based model that naturally figures out how to depict the substance of images. They describe how they can train this model in a deterministic way using standard back-propagation procedures and stochastically by amplifying a variational lower bound. At long last, (Zhu et al., 2018) used transformer model (Vaswani et al., 2017) for image captioning task, by feeding spatial visual features of images directly to transformer decoder, by using CNN encoder for images instead of transformer encoder for text.

With regards to NMT, (Dong et al., 2015) explored the issue of learning a machine translation model that can at the same time translate sentences from one source language to numerous target languages. They expanded the neural machine translation model to a multi-task learning structure which shares source language representation and isolates the modeling of various target language translation. Their structure can be connected to circumstances where either large amount of parallel data or restricted parallel data is accessible. (Firat et al., 2016) proposed a multi-way, multilingual neural machine translation. Their proposed approach empowers a solitary neural translation model to translate between numerous languages, with various parameters that becomes just directly with the quantity of languages. This is made conceivable by having a single attention mechanism that is shared over all language sets. (Luong et al., 2015) analyzed three multi-task learning settings for sequence to sequence models: 1) The one-to-many setting where the encoder is shared between a few tasks. 2) The many-to-one setting where just the decoder can be shared, as on account of translation and image captioning. 3) The many-to-many setting where various encoders and decoders are shared, which is the situation with unsupervised destinations and translation. In spite of the fact that it is not an NMT model, (Hitschler and Riezler, 2016) exhibited a way to deal with enhance statistical machine translation of image depictions by multimodal turns characterized in visual space. Their key thought is to perform image recovery over a database of images that are inscribed in the objective language, and use the captions of the most comparable images for crosslingual reranking of translation outputs. Their approach does not rely upon the accessibility of a lot of in-domain parallel data, however just depends on accessible huge datasets of monoligually captioned images, and on convolutional neural networks to figure image similarities.

Distinctive research groups have proposed to incorporate global and spatial visual features in re-ranking n-best lists created by an SMT framework or straightforwardly in an NMT structure (Caglayan et al., 2016; Calixto et al., 2016; Huang et al., 2016; Libovický et al., 2016; Shah et al., 2016). (Huang et al., 2016) proposed to use global visual features extracted with the VGG19 network (Simonyan and Zisserman, 2014) for a whole image, and furthermore for areas of the image acquired using the RCNN of (Girshick et al., 2013). (Calixto et al., 2017) proposed the first attention-based MNMT models, and *doubly-attentive model*, where two attention mechanisms are incorporated into one multimodal decoder, and the entire-model trained jointly end-to-end.

With our work, we propose the first doubly attentive transformer MNMT model, where two attention mechanisms are fused together under scaled dot-product attention level. Also, differently from the previous model, our model is able to train standalone image-captioning, language, machine translation, multi-modal machine translation models, or the mix of all these data types homogeneously, and this pretraining leads to significant improvement on MNMT translation quality.

## 3 Background and Notation

### 3.1 Transformer NMT

We described the transformer NMT model (Figure 1.) presented by (Vaswani et al., 2017) in this section. Given a source sequence $X = (x_1, x_2, ..., x_N)$ and its translation $Y = (y_1, y_2, ..., y_M)$, an NMT model means to construct a solitary neural network that translates X into Y by specifically learning to model $p(Y|X)$. The entire network consists of one *encoder* and one *decoder*. Each $x_i$ is a row index in a word embedding matrix $E_x \in \mathbb{R}^{|V_x| \times d_x}$, as well as each $y_j$ being an index in a target word embedding matrix $E_y \in \mathbb{R}^{|V_y| \times d_y}$, $V_x$ and $V_y$ are source and target vocabularies, and $d_x$ and $d_y$ are source and target word embeddings dimensionalities, respectively.

The encoder is made out of a heap of $N = 6$ identical layers. Each layer has two sub-layers. The first is a multi-head self-attention component, and the second is a basic, position-wise fully connected feed-forward network. There is a residual connection (He et al., 2015) around every one of the two sub-layers, trailed by layer normalization (Ba et al., 2016). That is, the output of each sub-layer is $LayerNorm(x + SubLayer(x))$, where $Sublayer(x)$ is simply the function actualized by the sub-layer itself. To encourage these residual connections, all sub-layers in the model, and additionally the embedding layers, create outputs of dimension $d_{model} = 512$.

The decoder is likewise made out of a heap of $N = 6$ identical layers. In addition to the two sub-layer in each encoder-layer, the decoder embeds a third sub-layer, which performs multi-head attention over the output of the encoder stack. Like the encoder, there are residual connections around every one of the sub-layers, trailed by layer normalization. There is a masking system in self-attention sub-layer in the decoder stack to keep positions from attending to consequent positions. This masking, joined with the fact that output embeddings are counterbalanced by one position, guarantees that the predictions for position $i$ can depend just on the known outputs at positions before $i$.

### 3.2 Multi-Head Attention

In transformer, the attention function is a query mapping and an arrangement of key-value sets to an output, where the query, keys, values, and output are all vectors. The output is processed as a weighted sum of the values, where the weight allocated to each value is computed by a compatibility function of the query with the corresponding key.

The particular attention in transformer is "Scaled Dot-Product Attention" (Figure 2.). The input involves queries and keys of dimension $d_k$, and values of dimension $d_v$. Each query is multiplied with all keys by dot product multiplication, and divided by $\sqrt{d_k}$, and softmax function is applied on output to gain the weights on the values.

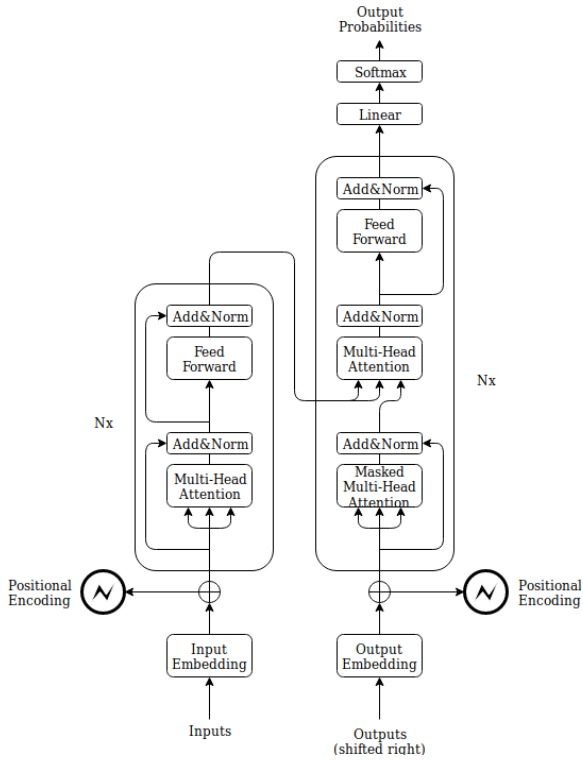

Figure 1: The Transformer - model architecture.

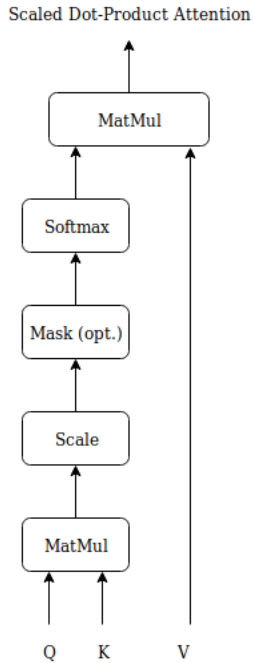

Figure 2: Scaled dot product attention architecture.

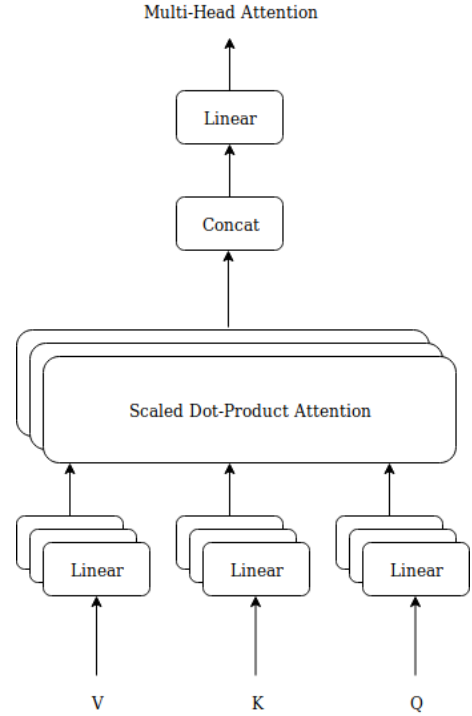

Figure 3: Multi-Head Attention architecture.

Practically, the attention function is simultaneously calculated on a set of queries, keys and values and packed together into a matrix $Q$, $K$ and $V$. The output matrix is computed as:

$$Attention(Q,K,V) = softmax(\frac{QK^T}{\sqrt{d_k}})V \quad (1)$$

With Multi-Head attention, rather than playing out a single attention work with $d_{model}$-dimensional keys, values and queries, these parameters are linearly projected $h$ times with various, learned linear projections to $d_k$, $d_k$ and $d_v$ dimensions, separately. On every one of these anticipated variants of queries, keys and values, the attention function is performed in parallel, yielding $d_v$-dimensional output values. The output values are projected and concatenated for the final values as in Figure 3.

Multi-head attention enables the model to mutu-ally take care of data from various representation subspaces at various positions. With a single attention head, averaging restrains this.

$$MultiHead(Q, K, V) = Concat(head_1, ..., head_h)W^O \quad (2)$$

where

$$head_i = Attention(QW_i^Q, KW_i^K, VW_i^V) \quad (3)$$

where the projections are parameter matrices

$$\begin{aligned} W_i^Q &\in \mathbb{R}^{d_{model} \times d_k}, \\ W_i^K &\in \mathbb{R}^{d_{model} \times d_k}, \\ W_i^V &\in \mathbb{R}^{d_{model} \times d_v}, \\ W^O &\in \mathbb{R}^{hd_v \times d_{model}} \end{aligned} \quad (4)$$

## 4 Doubly Attentive Transformer NMT

Our doubly attentive transformer NMT model (Figure 4.) can be viewed as an extension of the transformer NMT system depicted in §2.1.

We use openly accessible pre-trained CNNs for image feature extraction. In particular, we extract spatial image features for all images in our dataset using the 152-layer Residual network (ResNet-152) of (He et al., 2015).

These spatial feature are the activations of the res4f layer, which can be viewed as encoding an image in a 14x14 grid, where every one of the sections in the network is represented by a 1024D element vector that exclusive encodes data about that particular district of the image. We vectorise this 3-tensor into a 196x1024 network $A = (a_1, a_2, ..., a_L) \in \mathbb{R}^{1024}$ where every one of the $L = 196$ rows comprises of a 1024D element vector and every column represents one column in the image.

### 4.1 Doubly Attentive Transformer Decoder

Doubly attentive transformer integrates two separate attention mechanisms over the source-language words and visual features in a single transformer decoder. Our doubly-attentive transformer decoder fuses the scaled-dot production attention vectors of target words between encoder output features,

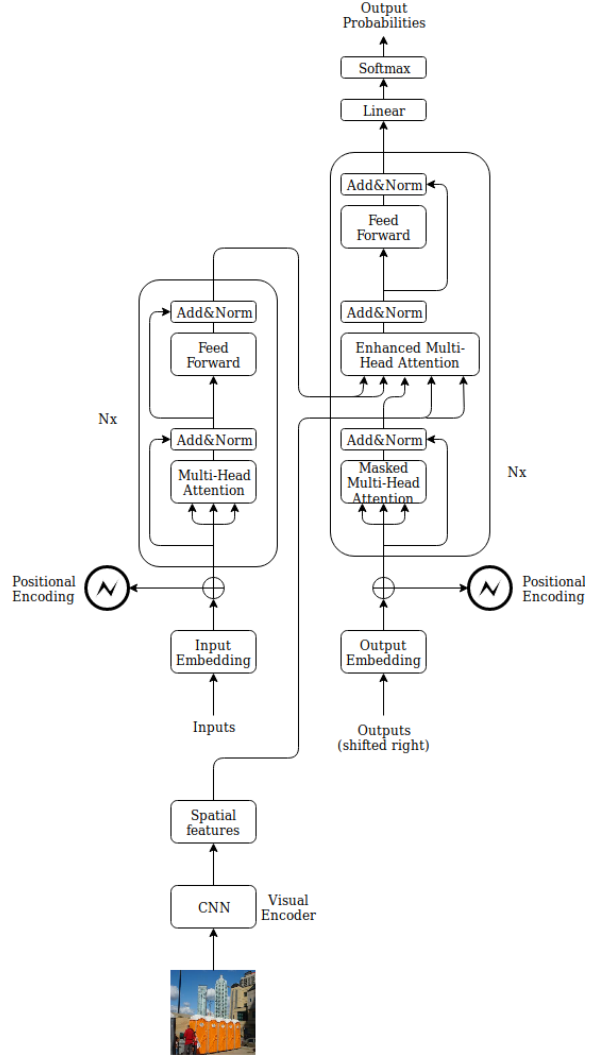

Figure 4: Doubly attentive Transformer NMT model architecture.

and between spatial visual features under Enhanced Multi-Head Attention layer.

$$\begin{aligned} Attention(Q, K_t, V_t, K_v, V_v) = \\ (softmax(\frac{QK_t^T}{\sqrt{d_k}})V_t) \\ + (softmax(\frac{QK_v^T}{\sqrt{d_k}})V_v) \end{aligned} \quad (5)$$

If attention to a target word at position $i$ is very strong from one of the sources (image, text), and very weak on another, summing two attention outputs as at (Equation 5.) still gives chance to the target word to have high probability to be learned by the model. In this way, if one of the sources does not have feature representing the target word, the model decides based on the other source.

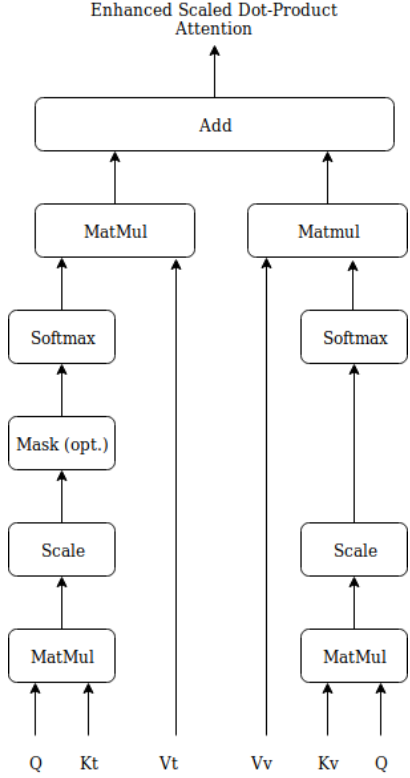

Figure 5: Enhanced scaled dot product attention architecture. Since there is no padding on spatial visual features vector, we don't use masking over visual vector.

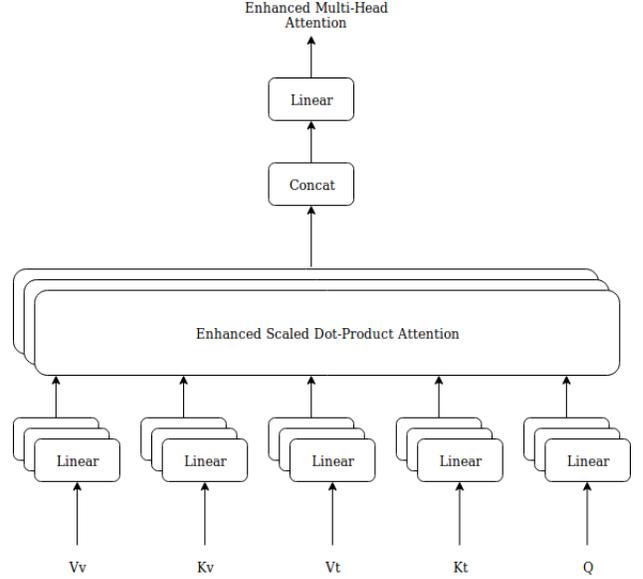

Figure 6: Enhanced Multi-Head attention architecture.

In this way, Enhanced Multi-Head Attention is represented as follows:

$$MultiHead(Q, K_t, V_t, K_v, V_v) = \\ Concat(head_1, ..., head_h)W^O \quad (6)$$

where

$$head_i = Attention(QW_i^Q, K_tW_i^{K_t}, V_tW_i^{V_t}, \\ K_vW_i^{K_v}, V_vW_i^{V_v},) \quad (7)$$

where the projections are parameter matrices

$$\begin{aligned}
W_i^Q &\in \mathbb{R}^{d_{model} \times d_k}, \\
W_i^{K_t} &\in \mathbb{R}^{d_{model} \times d_k}, \\
W_i^{V_t} &\in \mathbb{R}^{d_{model} \times d_v}, \\
W_i^{K_v} &\in \mathbb{R}^{d_{model} \times d_k}, \\
W_i^{V_v} &\in \mathbb{R}^{d_{model} \times d_v}, \\
W^O &\in \mathbb{R}^{hd_v \times d_{model}}
\end{aligned} \quad (8)$$

## 5 Data

The Flickr30K (Plummer et al., 2015) dataset contains 30k images and 5 descriptions in English and German for each image. In this work, we use Multi30k dataset (Elliott et al., 2016).

For every one of the 30k images in the Flickr30k, the M30k has one of the English depictions manually translated into German by an expert translator. Training, validation and test sets contain 29k, 1014 and 1000 pictures individually, each joined by a sentence match (the first English sentence and its translation into German).

We use the whole M30k training set for training our models, its validation set for model choice with BLEU (Papineni et al., 2002) and its test set for assessment. In addition, for pretraining models, we use whole English-German corpora from OPUS

Table 1: Datasets provided by OPUS. The corpora involves around 98 million sentence pairs, around 1.4 billion English and around 1.5 billion German tokens.

| Corpus | Sent's | En tokens | De tokens |
|---|---|---|---|
| ParaCrawl | 36.4M | 534.4M | 560.9M |
| EUbookshop (Skadiņš et al., 2014) | 9.6M | 337.4M | 380.2M |
| OpenSubtitles2018 (Lison and Tiedemann, 2016) | 24.4M | 176.3M | 191.3M |
| OpenSubtitles2016 (Lison and Tiedemann, 2016) | 15.4M | 110.5M | 118.8M |
| DGT (Tiedemann, 2012) | 3.2M | 55.4M | 72.9M |
| Europarl (Tiedemann, 2012) | 2.0M | 54.7M | 57.7M |
| Wikipedia (Tiedemann, 2012) | 2.5M | 43.5M | 58.4M |
| JRC-Acquis (Steinberger et al., 2006) | 0.7M | 30.7M | 34.1M |
| EMEA (Tiedemann, 2009) | 1.2M | 11.5M | 12.0M |
| Tanzil (Tiedemann, 2012) | 0.5M | 11.1M | 11.3M |
| News-Commentary11 (Tiedemann, 2012) | 0.2M | 6.5M | 6.6M |
| MultiUN (Tiedemann, 2012) | 0.2M | 5.7M | 6.2M |
| News-Commentary (Tiedemann, 2012) | 0.2M | 5.0M | 5.0M |
| GNOME (Tiedemann, 2012) | 0.8M | 4.4M | 5.5M |
| Tatoeba (Tiedemann, 2012) | 0.1M | 2.4M | 3.6M |
| ECB (Tiedemann, 2009) | 0.1M | 2.8M | 3.1M |
| TED2013 (Tiedemann, 2012) | 0.1M | 2.7M | 2.8M |
| KDE4 (Tiedemann, 2009) | 0.3M | 1.9M | 2.4M |
| GlobalVoices (Tiedemann, 2012) | 57.5k | 1.3M | 1.3M |
| Books (Tiedemann, 2012) | 52.3k | 1.1M | 1.3M |
| Ubuntu (Tiedemann, 2012) | 0.2M | 0.6M | 0.8M |
| OpenSubtitles (Tiedemann, 2009) | 76.0k | 0.5M | 0.6M |
| OpenOffice (Tiedemann, 2009) | 42.9k | 0.5M | 0.5M |
| WMT-News (Tiedemann, 2012) | 19.6k | 0.5M | 0.5M |
| PHP (Tiedemann, 2009) | 42.2k | 0.3M | 0.5M |
| EUconst (Tiedemann, 2009) | 9.0k | 0.2M | 0.2M |
| RF (Tiedemann, 2012) | 0.2k | 4.5k | 4.4k |
| **Total** | **98.5M** | **1.4G** | **1.5G** |

(Tiedemann and Nygaard, 2004) Table 1, and whole Flickr30k German descriptions.

We use tokenizer script in the Moses SMT Toolkit (Koehn et al., 2007) to tokenize English and German descriptions, and we also convert space-separated tokens into word-pieces with unigram model (Kudo, 2018) by using google sentencepiece model[1].

If the sentences in English or German are longer than 100 tokens, they are removed. We train models to translate from English to German and report assessment of cased, tokenized sentences with punctuation.

---

[1] https://github.com/google/sentencepiece

## 6 Experiments and Results

### 6.1 Experimental Setup

In our experiments, we mainly focus on the effect of incorporating visual features into the transformer NMT model, for this reason we use the same hyperparameter settings on original transformer work (Vaswani et al., 2017).

Visual features are extracted by feeding the images to the pretrained ResNet-152 and using the activations of the res4f layer (He et al., 2015). We apply linear projection on visual features to reduce dimension from 1024 to 512 to have the same size with word embeddings. We apply dropout of 0.5 on lin-ear projection.

We use sentence-minibatches of 32, where each training instance consists of one English sentence, one German sentence and one image for MNMT, one English sentence and one German sentence for NMT, and one image and one German sentence for image captioning. We use a homogeneous mixture of training instances for pretraining models. We train and save MNMT models for 100 epochs, and select the best model based on BLEU4 score on validation set. While, we select the best image captioning pretraining model based on best perplexity (Jelinek et al., 1977) on validation set, our NMT pretraining model is trained for 1 epoch. While the number of warmup steps in all the training setups is 4000, we see that this hyperparameter leads to overfitting for MNMT training with NMT pretrained model. In order to proceed finetuning with MNMT, we change this hyperparameter to 8000.

The translation quality of our models is assessed quantitatively with BLEU4 (Papineni et al., 2002), METEOR (Denkowski and Lavie, 2014) and TER (Snover et al., 2006). We report statistical significance with approximate randomisation for the three measurements using the MultEval tool (Clark et al., 2011).

### 6.2 Results and Discussions

In Table 2, we present results and comparisons, including Transformer NMT (TNMT), and our Doubly-Attentive Transformer NMT (DATNMT) results obtained without pretraining models. Tables 3 and 4 demonstrate the results of TNMT and DATNMT after pretraining models with Flickr30k, and large OPUS corpora, respectively.

**Training on M30k:** One primary finding is that our model reliably beats the equivalent model of NMT(SRC+IMG) (Calixto et al., 2017), with change of +4.5 BLEU. The experimental results show that while the text-only trained model with TNMT outperforms text-only trained model with NMT (RNN)(Sennrich et al., 2017) by +4.9 BLEU4 points, DATNMT outperforms TNMNT by +2.4 BLEU4 points. Moreover, Table 2 illustrates that NMT(SRC+IMG) performs better at recall-oriented metrics, i.e. METEOR and TER, whereas our model is better at precision-oriented ones, i.e. BLEU4.

**Pre-training:** We presently talk about outcomes for models pre-trained using distinctive data sets. First an image captioning model was pre-trained with our DATNMT on Flickr30k German data set, a medium-sized in-domain image description data set (145k training sentences). In addition, an NMT model on the English-German parallel sentences of much larger MT data sets on OPUS (Table 1) was pre-trained. From Table 2, it is clear that model DATNMNT can learn from both in-domain, image captioning data sets as well as text-only, general domain ones.

**Pre-training on Flickr30k:** On Table 3, When pre-training on Flickr30k, the precision-oriented BLEU4 shows a difference on TNMT. Although TNMNT does not use image information, the pre-trained language model on Flickr30k data set brings improvement to TNMT. However, the same improvement cannot be seen on DATNMT, although it has +0.1 BLEU point difference between TNMT.

**Pre-training on OPUS corpora:** In addition our models was pre-trained on OPUS corpora (Table 1) for 1 epoch, which took 10 days. Results show that DATNMT improves significantly over the TNMT baseline according to BLEU4, and is consistently the best according to all metrics. This is a strong indication that model DATNMT can exploit the additional pre-training data efficiently on general-domain text corpora.

**Textual and visual attention:** In Figure 7, the visual and textual attention weights for a sentence translation of the M30k test set are shown by using the **SoftAlignments** tool (Rikters et al., 2017). In the visual attention, it can be seen that when the words "Teenager" and "Trompete" are generated, the heatmap is focused on areas around the teenager's head. In addition, it can be observed that the target words which have strong attention with source words also shows strong attention on image parts.

Table 2: BLEU4, METEOR (higher is better) and TER scores (lower is better) on the translated Multi30k (M30k) test set without pretraining model. Best overall results appear in bold.

| Model | Training data | BLEU4 | METEOR | TER |
|---|---|---|---|---|
| NMT (RNN) | M30k | 33.7 | 52.3 | 46.7 |
| PBSMT | M30k | 32.9 | 54.3 | 45.1 |
| (Huang et al., 2016) | M30k | 35.1 | 52.2 | - |
| + RCNN | M30k | 36.5 | 54.1 | - |
| NMT(src+img) | M30k | 36.5 | **55.0** | **43.7** |
| TNMT | M30k | 38.6 | **55.0** | 51.6 |
| DATNMT | M30k | **41.0** | 53.5 | 48.4 |

Table 3: BLEU4, METEOR (higher is better) and TER scores (lower is better) on the translated Multi30k (M30k) test set after finetuning with image captioning pretraining model. Best overall results appear in bold.

| Model | Training data | BLEU4 | METEOR | TER |
|---|---|---|---|---|
| TNMT | M30k | 39.6 | **54.1** | **49.9** |
| DATNMT | M30k | **39.7** | 52.5 | 50.5 |

Table 4: BLEU4, METEOR (higher is better) and TER scores (lower is better) on the translated Multi30k (M30k) test set after finetuning with machine translation pre-training model. Best overall results appear in bold.

| Model | Training data | BLEU4 | METEOR | TER |
|---|---|---|---|---|
| TNMT | M30k | 41.3 | 57.1 | 48.4 |
| DATNMT | M30k | **42.2** | **57.2** | **45.9** |

# 7 Conclusion and Future Work

We have presented a novel transformer NMT model to fuse spatial visual data into NMT. We have reported that enhancing scaled-dot product attention layer with 2 attention vectors leads improvement on translation quality. We also pretrained our model homogeneous training sample types, and reported that our model is able to exploit general domain text-only or image captioning datasets for improving the translation accuracy.

In the future, we are planning to continue with multimodal image transformer, which will be the enhancement of image transformer (Parmar et al., 2018). We would like to see the affect of standalone text or speech, or their mixture on pixel-wise image generation.

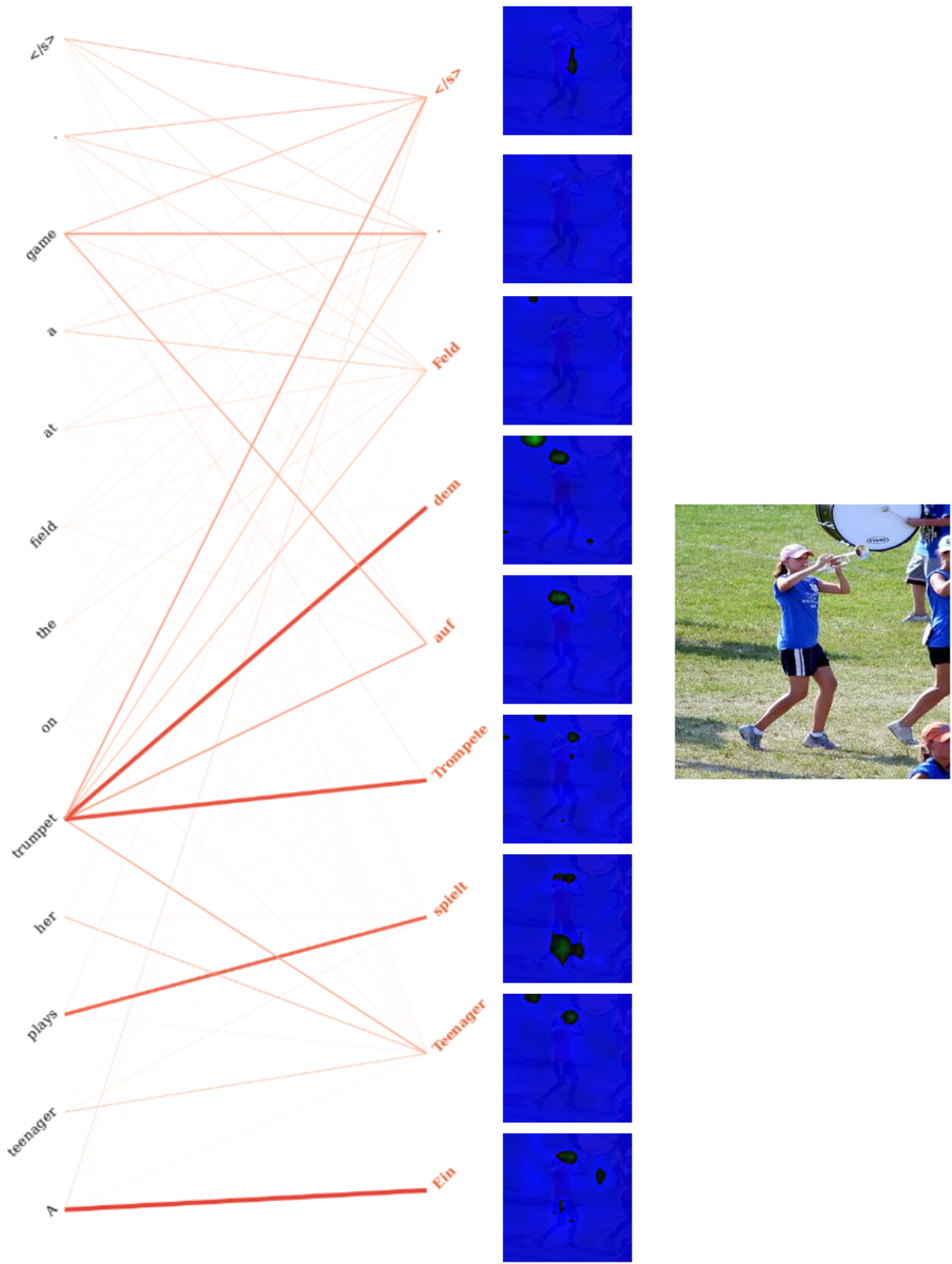

Figure 7: Textual attention map between source and target words and visual attention colormap between target words and subregions on the image for a translation example in Multi30k test set. The intense of the line color and spots on the image show the strength of the attention on text and image parts.

# References

Jimmy Lei Ba, Jamie Ryan Kiros, and Geoffrey E Hinton. 2016. Layer normalization. *arXiv preprint arXiv:1607.06450*.

Dzmitry Bahdanau, Kyunghyun Cho, and Yoshua Bengio. 2014. Neural machine translation by jointly learning to align and translate. *CoRR*, abs/1409.0473.

Ozan Caglayan, Walid Aransa, Yaxing Wang, Marc Masana, Mercedes García-Martínez, Fethi Bougares, Loïc Barrault, and Joost van de Weijer. 2016. Does multimodality help human and machine for translation and image captioning? *CoRR*, abs/1605.09186.

Iacer Calixto, Desmond Elliott, and Stella Frank. 2016. Dcu-uva multimodal mt system report. In *Proceedings of the First Conference on Machine Translation, WMT 2016, colocated with ACL 2016, August 11-12, Berlin, Germany*, pages 634–638. Association for Computational Linguistics (ACL), 8.

Iacer Calixto, Qun Liu, and Nick Campbell. 2017. Doubly-attentive decoder for multi-modal neural machine translation. *CoRR*, abs/1702.01287.

Kyunghyun Cho, Bart van Merrienboer, Çaglar Gülçehre, Fethi Bougares, Holger Schwenk, and Yoshua Bengio. 2014. Learning phrase representations using RNN encoder-decoder for statistical machine translation. *CoRR*, abs/1406.1078.

Jonathan H. Clark, Chris Dyer, Alon Lavie, and Noah A. Smith. 2011. Better hypothesis testing for statistical machine translation: Controlling for optimizer instability. In *Proceedings of the 49th Annual Meeting of the Association for Computational Linguistics: Human Language Technologies: Short Papers - Volume 2*, HLT '11, pages 176–181, Stroudsburg, PA, USA. Association for Computational Linguistics.

Michael Denkowski and Alon Lavie. 2014. Meteor universal: Language specific translation evaluation for any target language. In *In Proceedings of the Ninth Workshop on Statistical Machine Translation*.

Daxiang Dong, Hua Wu, Wei He, Dianhai Yu, and Haifeng Wang. 2015. Multi-task learning for multiple language translation. In *ACL*.

Desmond Elliott, Stella Frank, and Eva Hasler. 2015. Multi-language image description with neural sequence models. *CoRR*, abs/1510.04709.

Desmond Elliott, Stella Frank, Khalil Sima'an, and Lucia Specia. 2016. Multi30k: Multilingual english-german image descriptions. *CoRR*, abs/1605.00459.

Orhan Firat, KyungHyun Cho, and Yoshua Bengio. 2016. Multi-way, multilingual neural machine translation with a shared attention mechanism. *CoRR*, abs/1601.01073.

Ross B. Girshick, Jeff Donahue, Trevor Darrell, and Jitendra Malik. 2013. Rich feature hierarchies for accurate object detection and semantic segmentation. *CoRR*, abs/1311.2524.

Kaiming He, Xiangyu Zhang, Shaoqing Ren, and Jian Sun. 2015. Deep residual learning for image recognition. *CoRR*, abs/1512.03385.

Julian Hitschler and Stefan Riezler. 2016. Multimodal pivots for image caption translation. *CoRR*, abs/1601.03916.

Po-Yao Huang, Frederick Liu, Sz-Rung Shiang, Jean Oh, and Chris Dyer. 2016. Attention-based multimodal neural machine translation. In *WMT*.

F. Jelinek, R. L. Mercer, L. R. Bahl, and J. K. Baker. 1977. Perplexity – a measure of the difficulty of speech recognition tasks. *Journal of the Acoustical Society of America*, 62:S63, November. Supplement 1.

Nal Kalchbrenner and Phil Blunsom. 2013. Recurrent continuous translation models. In *Proceedings of the 2013 Conference on Empirical Methods in Natural Language Processing*, pages 1700–1709. Association for Computational Linguistics.

Philipp Koehn, Hieu Hoang, Alexandra Birch, Chris Callison-Burch, Marcello Federico, Nicola Bertoldi, Brooke Cowan, Wade Shen, Christine Moran, Richard Zens, Chris Dyer, Ondřej Bojar, Alexandra Constantin, and Evan Herbst. 2007. Moses: Open source toolkit for statistical machine translation. In *Proceedings of the 45th Annual Meeting of the ACL on Interactive Poster and Demonstration Sessions*, ACL '07, pages 177–180, Stroudsburg, PA, USA. Association for Computational Linguistics.

Taku Kudo. 2018. Subword regularization: Improving neural network translation models with multiple subword candidates. *CoRR*, abs/1804.10959.

Jindrich Libovický, Jindrich Helcl, Marek Tlustý, Pavel Pecina, and Ondrej Bojar. 2016. CUNI system for WMT16 automatic post-editing and multimodal translation tasks. *CoRR*, abs/1606.07481.

Pierre Lison and Jörg Tiedemann. 2016. Opensubtitles2016: Extracting large parallel corpora from movie and tv subtitles.

Minh-Thang Luong, Quoc V. Le, Ilya Sutskever, Oriol Vinyals, and Lukasz Kaiser. 2015. Multi-task sequence to sequence learning. *CoRR*, abs/1511.06114.

Junhua Mao, Wei Xu, Yi Yang, Jiang Wang, and Alan L. Yuille. 2014. Explain images with multimodal recurrent neural networks. *CoRR*, abs/1410.1090.

Kishore Papineni, Salim Roukos, Todd Ward, and Wei-Jing Zhu. 2002. Bleu: A method for automatic evaluation of machine translation. In *Proceedings of the 40th Annual Meeting on Association for Computational Linguistics*, ACL '02, pages 311–318, Stroudsburg, PA, USA. Association for Computational Linguistics.

Niki Parmar, Ashish Vaswani, Jakob Uszkoreit, Lukasz Kaiser, Noam Shazeer, and Alexander Ku. 2018. Image transformer. *CoRR*, abs/1802.05751.

Bryan A. Plummer, Liwei Wang, Chris M. Cervantes, Juan C. Caicedo, Julia Hockenmaier, and Svetlana Lazebnik. 2015. Flickr30k entities: Collecting region-to-phrase correspondences for richer image-to-sentence models. *CoRR*, abs/1505.04870.

Matīss Rikters, Mark Fishel, and Ondřej Bojar. 2017. Visualizing neural machine translation attention and confidence. *The Prague Bulletin of Mathematical Linguistics*, 109(1):39–50.

Rico Sennrich, Orhan Firat, Kyunghyun Cho, Alexandra Birch, Barry Haddow, Julian Hitschler, Marcin Junczys-Dowmunt, Samuel Läubli, Antonio Valerio Miceli Barone, Jozef Mokry, et al. 2017. Nematus: a toolkit for neural machine translation. *arXiv preprint arXiv:1703.04357*.

Kashif Shah, Josiah Wang, and Lucia Specia. 2016. Shef-multimodal: Grounding machine translation on images. In *WMT*.

Karen Simonyan and Andrew Zisserman. 2014. Very deep convolutional networks for large-scale image recognition. *CoRR*, abs/1409.1556.

Raivis Skadiņš, Jörg Tiedemann, Roberts Rozis, and Daiga Deksne. 2014. Billions of parallel words for free: Building and using the eu bookshop corpus. In *Proceedings of the 9th International Conference on Language Resources and Evaluation (LREC-2014)*, Reykjavik, Iceland, May. European Language Resources Association (ELRA).

Matthew Snover, Bonnie J. Dorr, Richard F. Schwartz, and Linnea Micciulla. 2006. A study of translation edit rate with targeted human annotation.

Ralf Steinberger, Bruno Pouliquen, Anna Widiger, Camelia Ignat, Tomaz Erjavec, Dan Tufis, and Dániel Varga. 2006. The jrc-acquis: A multilingual aligned parallel corpus with 20+ languages. *arXiv preprint cs/0609058*.

Ilya Sutskever, Oriol Vinyals, and Quoc V. Le. 2014. Sequence to sequence learning with neural networks. *CoRR*, abs/1409.3215.

Jörg Tiedemann and Lars Nygaard. 2004. The opus corpus-parallel and free: http://logos. uio. no/opus. In *LREC*. Citeseer.

Jörg Tiedemann. 2009. News from OPUS - A collection of multilingual parallel corpora with tools and interfaces. In N. Nicolov, K. Bontcheva, G. Angelova, and R. Mitkov, editors, *Recent Advances in Natural Language Processing*, volume V, pages 237–248. John Benjamins, Amsterdam/Philadelphia, Borovets, Bulgaria.

Jörg Tiedemann. 2012. Parallel data, tools and interfaces in opus. In Nicoletta Calzolari (Conference Chair), Khalid Choukri, Thierry Declerck, Mehmet Ugur Dogan, Bente Maegaard, Joseph Mariani, Jan Odijk, and Stelios Piperidis, editors, *Proceedings of the Eight International Conference on Language Resources and Evaluation (LREC'12)*, Istanbul, Turkey, may. European Language Resources Association (ELRA).

Ashish Vaswani, Noam Shazeer, Niki Parmar, Jakob Uszkoreit, Llion Jones, Aidan N. Gomez, Lukasz Kaiser, and Illia Polosukhin. 2017. Attention is all you need. *CoRR*, abs/1706.03762.

Subhashini Venugopalan, Marcus Rohrbach, Jeff Donahue, Raymond J. Mooney, Trevor Darrell, and Kate Saenko. 2015. Sequence to sequence - video to text. *CoRR*, abs/1505.00487.

Oriol Vinyals, Alexander Toshev, Samy Bengio, and Dumitru Erhan. 2014. Show and tell: A neural image caption generator. *CoRR*, abs/1411.4555.

Kelvin Xu, Jimmy Ba, Ryan Kiros, Kyunghyun Cho, Aaron C. Courville, Ruslan Salakhutdinov, Richard S. Zemel, and Yoshua Bengio. 2015. Show, attend and tell: Neural image caption generation with visual attention. *CoRR*, abs/1502.03044.

Xinxin Zhu, Lixiang Li, Jing Liu, Haipeng Peng, and Xinxin Niu. 2018. Captioning transformer with stacked attention modules. *Applied Sciences*, 8(5):739.